\pdfoutput=1

\documentclass[11pt]{article}

\usepackage[]{acl}

\usepackage{times}
\usepackage{latexsym}

\usepackage[T1]{fontenc}

\usepackage[utf8]{inputenc}

\usepackage{microtype}

\usepackage{amsfonts,amsmath}
\usepackage{mathrsfs}
\usepackage{graphicx}
\graphicspath{ {./figures/} }
\usepackage{caption,subcaption}
\usepackage{booktabs}
\usepackage{color}
\usepackage{xspace}
\usepackage{multirow,enumitem}

\DeclareMathOperator*{\argmaxk}{argmax_K}

\newcommand{\secref}[1]{Section~\ref{#1}\xspace}
\newcommand{\tabref}[1]{Table~\ref{#1}\xspace}

\newcommand{\set}[1]{\mathcal{#1}\xspace}
\newcommand{\dataset}[1]{\texttt{#1}\xspace}
\newcommand{\model}[1]{\textsc{#1}\xspace}
\usepackage[normalem]{ulem}

\definecolor{applegreen}{rgb}{0.55, 0.71, 0.0}

\title{Pan More Gold from the Sand: Refining Open-domain Dialogue Training with Noisy Self-Retrieval Generation}

\author{
Yihe Wang\textsuperscript{\rm 1}\thanks{ \ \ Work done during internship at Noah’s Ark Lab, Huawei} , Yitong Li\textsuperscript{\rm 2,3}\thanks{ \ \ Equal Contribution} ,  Yasheng Wang\textsuperscript{\rm 2}, Fei Mi\textsuperscript{\rm 2} \\ \textbf{Pingyi Zhou}\textsuperscript{\rm 2}, \textbf{Xin Wang}\textsuperscript{\rm 1}, \textbf{Jin Liu}\textsuperscript{\rm 1}\thanks{ \ \ Corresponding Author} , \textbf{Xin Jiang}\textsuperscript{\rm 2}, \textbf{Qun Liu}\textsuperscript{\rm 2} \\
\textsuperscript{\rm 1}School of Computer Science, Wuhan University \\ \textsuperscript{\rm 2}Noah’s Ark Lab, Huawei \quad \textsuperscript{\rm 3}Huawei Technologies Ltd.
\\{\small \texttt \{yihewang, jinliu, xinwang0920\}@whu.edu.cn} 
\\{\small \texttt \{liyitong3, wangyasheng, feimi2, zhoupingyi, Jiang.Xin, qun.liu\}@huawei.com} 
}

\begin{document}
\maketitle
\begin{abstract}
Real human conversation data are complicated, heterogeneous, and noisy, from which building open-domain dialogue systems remains a challenging task.
In fact, such dialogue data still contains a wealth of information and knowledge, however, they are not fully explored.
In this paper, we show existing open-domain dialogue generation methods that memorize \textit{context-response} paired data with autoregressive or encode-decode language models underutilize the training data.
Different from current approaches, using external knowledge, we explore a retrieval-generation training framework that can take advantage of the heterogeneous and noisy training data by considering them as "evidence".
In particular, we use BERTScore for retrieval, which gives better qualities of the evidence and generation.
Experiments over publicly available datasets demonstrate that our method can help models generate better responses, even such training data are usually impressed as low-quality data.
Such performance gain is comparable with those improved by enlarging the training set, even better.
We also found that the model performance has a positive correlation with the relevance of the retrieved evidence.
Moreover, our method performed well on zero-shot experiments, which indicates that our method can be more robust to real-world data.
\end{abstract}

\section{Introduction}

Open-domain dialogue is a long-standing problem in natural language processing and has aroused the widespread interest of researchers. Many approaches have been studied, and recently, generation models trained on large-scale data have gained more attention \cite{adiwardana2020towards,roller2020recipes,xu2021beyond,madotto2021few,bao2019plato,bao2020plato,zhang2019dialogpt,wang2020chinese}.
Open-domain dialogue systems are born to deal with diverse domains, and naturally their training data, usually crawled from online resources such as Reddit and Twitter, are heterogeneous and contain utterances with many various topics, more freedom of topic shifting, and vague responses \cite{kummerfeld2018large}.
As a result, directly building generation models from such data will be inefficient and usually requires "knowledge" during the training.

One common solution is to introduce external knowledge, usually, in a form of unstructured knowledge passages from Wikipedia \cite{dinan2018wizard} or Internet articles \cite{komeili2021internet},
and then, to build retrieval-augmented generation (RAG) methods to improve the response quality \cite{lewis2020retrieval,izacard2020leveraging}.
However, this assumes knowledge-intensive scenarios, which are not suitable for general open-domain or robust to noise.
According to our preliminary study, in the Reddit dataset, 43\% of the dialogues are merely chitchat and cannot match "knowledge".
Moreover, building such a knowledge-augmented dataset is very expensive as it relies on large amounts of high-quality human annotations w.r.t. knowledge grounding.
And thus, they are limited in size, making it hard for a knowledge-retrieval method to generalize on scale.

Motivated by the above, we would like to investigate \textit{can we have better ways of utilizing open domain data without introducing external resources?}
To tackle the aforementioned problem, we found that the context from the other relevant dialogue sessions can still be very useful for dialogue generation.
To utilize such unstructured contexts, we take inspiration from retrieval-augmented methods \cite{lewis2020retrieval}.
Differently, we retrieve useful dialogue context as evidence, build context-evidence-response triples for each dialogue turn, and treat open-domain generation as an evidence-aware generation task.
Such that our model can learn to respond with useful grounding evidences.
To retrieve evidences, we adopt similarity-based BERTScore~\cite{zhang2019bertscore}, which leverages pre-trained contextual embeddings from BERT and matches words in two sentences by cosine similarity.
It has been shown to correlate with human judgment on sentence-level and system-level evaluation. Although it was proposed as an automatic evaluation metric for text generation, due to the high correlation with human judgment, we consider it as a better off-the-shelf method to pick high-relevant evidences, compared with lexicon-based BM25. 

By this, we show that current training methods which learn merely using context-response pairs have not fully unleashed the potential of training data and that our methods, only retrieving from the training data, can consistently improve the generation performance.
We also perform zero-shot experiments, demonstrating that our method can be robust and generalized to different domains.
Moreover, we found that adding extra retrieval data only (without training them) can still help the model gain performance, and it can even outperform traditional methods directly trained on that part of retrieval data.
This proves our method is compatible with current methods with external knowledge.

Our contributions are summarized as follows:
\begin{itemize}[itemsep=0pt,topsep=2pt,leftmargin=12pt]
\item we explore a retrieval-generation training framework that can increase the usage of training data by directly considering the heterogeneous and noisy training data as the "evidence".
\item We show that adding extra retrieval data while not training them can still gain performance benefits, even better than traditional training with the retrieval data attached.
\item The proposed method performs well on zero-shot experiments, which indicates that our method can generalize well in real-world applications.
\end{itemize}

\section{Related Work}
\paragraph{Open-domain Dialogue System} Open-domain dialogue system aims to perform chit-chat with people without the task and domain restriction.
\citet{adiwardana2020towards} proposed Meena, a multi-turn open-domain chatbot trained end-to-end on data mined and filtered from public domain social media conversations. Blender \cite{roller2020recipes,xu2021beyond} learn to provide engaging talking points and listen to their partners, as well as display knowledge, empathy and personality appropriately, while maintaining a consistent persona. Adapter-bot \cite{madotto2021few} explored prompt-based few-shot learning in dialogue tasks. Plato \cite{bao2019plato,bao2020plato} introduced discrete latent variables to tackle the inherent one-to-many mapping problem in response generation.  \citet{zhang2019dialogpt} proposed DialoGPT which was trained on 147M conversation-like exchanges extracted from Reddit comment chains. \citet{wang2020chinese} introduced CDial-GPT, a pre-training dialogue model which is trained on a large-scale cleaned Chinese conversation dataset. \citet{mi2022pangubot} built PANGU-BOT with relatively fewer data and computation costs by inheriting valuable language capabilities and knowledge from pre-trained language model.

\begin{figure*}[!t]
\centering
\includegraphics[width=0.95\textwidth]{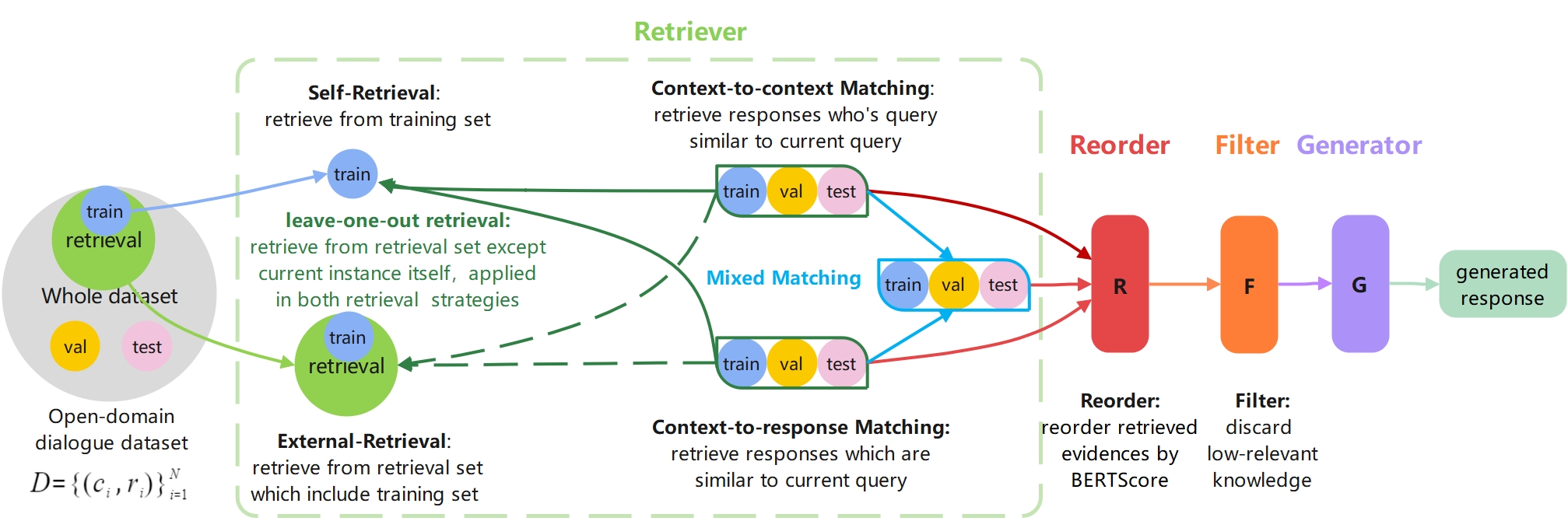} 
\caption{Overview of our self-retrieval approach as well as external-retrieval approach. In self-retrieval, our retriever first retrieves useful dialogue instances from the training dataset, which extends current data to context-evidence-response triples. And then, we adopt evidence-aware training models over the data with self-retrieval evidences.}
\label{figure:model}
\vspace{-0.2in}
\end{figure*}

\paragraph{Retrieval Augmented Generation}
Retrieval is a long-considered intermediate step in dialogue systems, and recently, it has been an intensively studied topic for neural models\cite{songensemble, pandey2018exemplar, weston2018retrieve, wu2019response, cai2019skeleton}.
\citeauthor{lewis2020retrieval} explored a fine-tuning recipe for retrieval-augmented generation, which combined pre-trained parametric and non-parametric memory for language generation. \citeauthor{izacard2020leveraging} proposed Fusion-in-Decoder which encoded each evidence independently with the context when generative model processing retrieved passages. \citet{li2022exploring} explored how to effectively utilize information with different channel settings of FiD in multi-turn topic driven Conversations. 
Most of these works retrieved external knowledge, usually unstructured knowledge passages, such as Wizard of Wikipedia \cite{dinan2018wizard}, persona-chat \cite{zhang2018personalizing}, and Wizard of Internet \cite{komeili2021internet}.
Moreover, \citet{li2020zero} proposed a zero-resource knowledge-grounded dialogue model which bridged a context and a response as knowledge and expressed it as a latent variable.

\section{Self-retrieval Method}

We start from an open-domain dialogue dataset $\set{D} = \{(c_i, r_i)\}_{i=1}^N $, where $c_i$ denotes multi-turn dialogue context, consisting of dialogue utterances, and $r_i$ represents the response.

Generally, we aim to build open-domain dialogue systems that retrieve useful dialogue responses (as evidences) from other sessions to help response generation.
To tackle this problem, we proposed a two-step framework.
The overview of our approach is shown in Figure~\ref{figure:model}.
\begin{enumerate}[itemsep=0pt,topsep=2pt,leftmargin=12pt]
\item Firstly, we extend an open-domain dialogue dataset with a \textit{retriever}.
Given the context of current dialogue turn $c_i$, the retriever $\mathscr{R}(e_{\{\cdot\}}|c_i)$ returns top-$k$ relevant evidences as the \textit{evidence set} $\set{E}_i = \{e_{1:k}\}$ from a \textit{retrieval set}.
Note that different from existing knowledge-grounding methods, we do not introduce external data for our retriever, and we only consider retrieving evidence from the training data at hand.
By that, we extend the dataset into context-evidence-response triples $\set{D} = \{(c_i, \set{E}_i, r_i)\}_{i=1}^N$.

\item Secondly, we adopt an evidence-aware generation model, which is a conditional language model to generate the response $y$ given the context and the retrieved evidence $p(y| c, \set{E})$.
We investigate two widely used architectures, an auto-regressive GPT, and an encoder-decoder based language model T5.
\end{enumerate}

Next, we introduce how to design an effective retriever in \secref{subsec:retriever} and ways of implementing evidence-aware generation on the basis of state-of-the-art pre-trained generation models in \secref{subsec:generation}.

\subsection{Retrieve Dialogue Evidence}
\label{subsec:retriever}

A variety of retrieval systems have been studied, including classic but effective bag-of-words system \cite{robertson1995okapi} and up-to-date dense retriever, such as DPR \cite{karpukhin2020dense} and SparTerm \cite{bai2020sparterm}.
We utilized an off-the-shelf similarity based \textbf{BERTScore} to retrieve evidence \cite{zhang2019bertscore}.\footnote{We also did preliminary experiments over BM25 and it shows no significant differences for our findings.}
BERTScore computes token similarity using pre-trained contextual embeddings rather than exact matches, which shows better coherent matching capability compared with human judgment.
During the retrieval, for each context-response pair $(c_i, r_i)$, we define the \textit{retrieval set} by applying leave-one-out of the original training set $\set{S} = \set{D} - \{(c_i, r_i)\}$, to ensure the model cannot see the true response during generation.


We explore three retrieval strategies: context-to-context (\model{c2c}) retrieval, context-to-response (\model{c2r}) retrieval, and a \model{mix} retrieval.

\paragraph{Context-to-context Matching}
\model{c2c} matches the context $c_i$ of current dialogue and the context $c_j$ from the retrieval set $\set{S}$.
And the evidence set of $c_i$ is defined as:
\begin{align*}
    \set{E}_i^{\model{c2c}}(c_i, \set{S}) = \argmaxk_{(c_j, r_j) \in \set{S}} \operatorname{score}(c_i, c_j) \, ,
\end{align*}
where $\argmaxk$ means selecting top $k$ corresponding responses $r_{1:k}$ as evidences $e_{1:k}$ with best matching $\operatorname{score}$ given by \model{BERTScore}.

\paragraph{Context-to-response Matching}
As the retrieval set contains the dialogue response, we also perform a Context-to-response (\model{c2r}) Matching.
It is similar to \model{c2c}, while \model{c2r} directly matches the response in the retrieval set.
In \model{c2r}, \model{BERTScore} computes the matching score based on the response $r_j$ of the retrieval set.
\begin{align*}
    \set{E}_i^{\model{c2r}}(c_i, \set{S}) = \argmaxk_{(c_j, r_j) \in \set{S}} \operatorname{score}(c_i, r_j) \, .\vspace{-0.2in}
\end{align*}

\paragraph{Mixed Matching}
We observed that these two strategies, \model{c2c} and \model{c2r}, often obtain different results.
Therefore, we complement the two retrieval sets of \model{c2c} and \model{c2r} with each other and combine them into a \model{mix} retrieval set by re-ranking them using BERTScore.
Finally, we take their responses as evidences:
\begin{align*}
    \set{E}_i^{\model{mix}}(u_i, \set{S}) = \argmaxk \{ \set{E}_i^{\model{c2c}}, \set{E}_i^{\model{c2r}} \} \, .
\end{align*}
\paragraph{Filter}
During preliminary studies, we found that some retrieved evidences are not relevant to the current context.
It is arguable that very few relevant evidences can be retrieved for some dialogue instances, and to study this we perform analysis in \secref{sec:filter}, where we study different sizes of the retrieval set to ensure more relevant evidences can be found.
Undoubtedly, these low-relevant evidences are harmful to response generation.
Therefore, we approach a simple filter to discard evidences with very low matching scores.

\subsection{Evidence-aware Dialogue Generation}
\label{subsec:generation}
For generating more appropriate responses, our generator is a language model but also conditional on the retrieved evidence set.\footnote{Note that responses from the retrieval set are not directly trained by the language model, but used as the evidences at the input side only.}
\begin{equation*}
    p(y | c_i, \set{E}_i) = \prod_{t} p(y_t | c_i, \set{E}_i, y_{<t}) \, .
\end{equation*}
Generally speaking, it can be modeled by any auto-regressive or encoder-decoder generation architecture for open-domain dialogue.
To demonstrate, we adopt both widely used architectures, i.e. a \model{GPT-2} \cite{Brown2020LanguageMA} and a Fusion-in-Decoder (\model{FiD}; \citealp{izacard2020leveraging}).\footnote{We also experiment with T5 architectures via concatenating the context and evidences and decoding the response. Yet the performance does not significantly vary from GPT thus we do not report \model{T5} in our main results.}

\paragraph{GPT-2}
GPT \cite{radford2019language,Brown2020LanguageMA} is auto-regressive language model based on multi-head self-attention transformers \cite{vaswani2017attention}.
For our task, the model takes the dialogue context and the support evidences as the input, and then it generates the response.
More precisely, for any instance $(c_i, \set{E}_i, r_i)$, all retrieved evidences are concatenated before the dialogue context $c_i$, and the model directly generates the response $y$ after $c_i$.
We add special token \texttt{[p]} before each retrieved evidence passage, and following \citet{wang2020chinese}, we add \texttt{[speaker1]}, \texttt{[speaker2]} to each utterance to indicate different speakers of muti-turn dialogue.

\paragraph{Fusion-in-Decoder}
In our setups, we have multiple evidences for one instance, thus we adopt a slightly different model than the standard encoder-decoder \model{T5} \cite{JMLR:v21:20-074}.
We use \model{FiD} \cite{izacard2020leveraging}, which was originally proposed for open-domain question answering.
It considers encoding each evidence independently with context, so that these evidences will not affect each other on the encoder side, which is a better solution to encode multiple evidences.
In detail, \model{FiD} encodes a concatenation of the context $c_i$ with each retrieved evidence $e_j$.
It concatenates all the encoded hidden representations and then passed to the decoder for response generation.
Slightly different from the original architecture, we add an additional passage that only encodes the dialogue context, in case one dialogue does not use any retrieved evidence (discussed in \secref{sec:relev}).
Similarly, we add special tokens as we did for {\model{GPT-2}}.

\section{Experiments}

\subsection{Datasets}

To evaluate the performance of the proposed model, we conduct experiments on two publicly available dialogue datasets.
\paragraph{Reddit Dataset}
The \dataset{Reddit} dataset is extracted from comment chains scraped from Reddit spanning.
Reddit discussions can be naturally expanded as tree-structured reply chains, since a thread replying to one thread forms the root node of subsequent threads.
We derived the dataset from DialoGPT \cite{zhang2019dialogpt}, and use their script to obtain and process the full dataset or demo dataset.\footnote{\url{https://github.com/microsoft/DialoGPT}.}
We report results on the demo dataset which comprises 770k multi-turn dialogue instances and is sufficient for our experiments.

\paragraph{Movie Dialog Dataset}
\dataset{Movie dialog} dataset collects movie discussions from real conversation taken directly under the \textit{movie} subreddit \cite{dodge2015evaluating}.\footnote{\url{https://research.fb.com/downloads/babi/}.}
We discard instances with long turns or long sentences.
In total, the movie dialog dataset has $940k$ dialogue sessions after preprocessing.

\begin{table*}[!t]
\centering
\small
\begin{tabular}{lcccccc|cccc}
\toprule
 & & \multicolumn{5}{c}{Automatic Metrics} & \multicolumn{3}{c}{Human Evaluation}\\
 \cmidrule(lr){3-7} \cmidrule(lr){8-11}
 
\dataset{Reddit} & & \textbf{PPL}$\downarrow$ & \textbf{F1}$\uparrow$ & \textbf{BLEU}$\uparrow$ & \textbf{Dist-1}$\uparrow$ & \textbf{Dist-2}$\uparrow$ & \textbf{Flue}$\uparrow$ & \textbf{Info}$\uparrow$ & \textbf{Relv}$\uparrow$ & \textbf{SSA}$\uparrow$ \\

\midrule
\model{GPT-2} & \model{baseline} & 31.3 &5.3 &3.4  &65.4 &96.7  &3.0 	 &2.9 	 &2.8 &46\% \\
\multirow{4}{*}{w. \model{SR}}
& BM25 \model{mix}  &28.1 &6.6 	&4.2 &73.5 	&\textbf{98.2} &3.4  &3.3 	 &3.4 &51\%\\
&BERTScore \model{c2c}  &27.7 &7.2 	&4.8 	&75.2 	&96.1 &3.4  &3.4 	 &3.4 &54\%\\
  &BERTScore \model{c2r}  &27.9 	&7.0 	&4.7 	&75.0	&96.4  &3.3 	 &3.4 	 &3.3 &53\%\\
   &BERTScore \model{mix}  &\textbf{27.1} &\textbf{7.8}	&\textbf{5.4}	&\textbf{76.1}	&96.8  	&\textbf{3.5}	&\textbf{3.4}	&\textbf{3.5} &\textbf{55\%}\\
\midrule

\model{T5}  &\model{baseline} & 25.5 	&5.2	&3.7 	&\textbf{95.7}	&96.3 &3.1 &3.0 &3.1 &48\%\\
\multirow{4}{*}{\model{FiD} w. \model{SR}} 
& BM25 \model{mix}  &23.8 &9.5 	&6.9 	&95.3 	&\textbf{97.2} &3.5 &3.4  &3.5 &52\%\\
 &BERTScore \model{c2c}  &23.3	&9.9 	&7.3	&94.3 	&94.7 &3.5 &3.5 &3.4 &54\%\\
 &BERTScore \model{c2r}   &23.4	&9.8 	&7.2 	&94.0 	&94.4 &3.5 &3.4 &3.4 &54\%\\
 &BERTScore \model{mix}   &\textbf{22.7}	&\textbf{10.4}	&\textbf{7.8} 	&95.6	&96.5 &\textbf{3.6} &\textbf{3.5} &\textbf{3.5} &\textbf{56\%}\\
\midrule \midrule

\dataset{Movie} & & \textbf{PPL} & \textbf{F1} & \textbf{BLEU} & \textbf{Dist-1} & \textbf{Dist-2} & \textbf{Flue} & \textbf{Info} & \textbf{Relv} & \textbf{SSA}\\

\midrule
\model{GPT-2}  & \model{baseline} & 25.6 &5.4 &3.3 &64.3 &96.0 &3.0 	&2.9 	&2.8 &47\%\\
\multirow{4}{*}{w. \model{SR}} 
& BM25 \model{mix}  &22.7 &6.7 	&4.2 	&71.4 	&96.1 &3.4 &3.3  &3.3 &52\%\\
 &BERTScore \model{c2c}  &22.1 &7.1 &4.7 &71.7 &94.9 &3.4 	&3.4 	&3.3 &53\%\\
 &BERTScore \model{c2r}  &22.3 &7.0 &4.7 &72.0 &94.3  &3.3 	&3.4 	&3.3 &53\%\\
 &BERTScore \model{mix}  &\textbf{21.6} 	&\textbf{7.6}	&\textbf{5.2}	&\textbf{73.4}	&\textbf{96.2} &\textbf{3.5}	&\textbf{3.4} 	&\textbf{3.4} &\textbf{55\%}\\
\midrule
\model{T5}  & \model{baseline}	&20.5	&5.2	&3.7  &95.2 	&95.8 &3.1 	&2.9 &	2.9  &48\%\\
\multirow{4}{*}{\model{FiD} w. \model{SR}}
 & BM25 \model{mix}  &18.9	&9.2 &6.6  &94.9 &96.8 &3.6 &3.5 &3.6 &53\%\\
 &BERTScore \model{c2c}  &18.4	&9.5 &7.0  &94.4 &95.6 &3.6 &3.6 &3.5 &55\%\\
 &BERTScore \model{c2r}  &18.5	&9.4	&6.8  &93.8 	&94.9 &3.5 	&3.5 	&3.6 &54\%\\
 &BERTScore \model{mix}  &\textbf{17.9}	&\textbf{10.1}	&\textbf{7.5}  &\textbf{95.3}	&\textbf{96.9} &\textbf{3.7}	&\textbf{3.7}	&\textbf{3.6} &\textbf{57\%}\\
\bottomrule
\end{tabular}
\caption{
 Automatic and human evaluation of the in-domain setups over \dataset{Reddit} and \dataset{Movie Dialog}, using 8 evidences passages. \model{GPT-2} and \model{T5} are baselines.  ``w. \model{SR}'' (with self-retrieval) indicate our methods. The best results are in \textbf{bold}. 
}
\label{table: overview}
\end{table*}

For both datasets, we randomly sample a training set of 100k samples, a validation set of 10k samples, and a test set of 10k samples.
Data outside the above sets can be considered as retrieval resources.
Noted that in our main experiments, the retrieval set (for train/dev/test) is exactly the training set, where we only retrieve from the training set.
And experimental results using a larger retrieval set are investigated and reported in \secref{sec:extra}, which involves more evidence than the training set.

\subsection{Metrics}
To evaluate response quality, we adopt both automatic metrics and human evaluations.

\paragraph{Automatic Metrics} We deploy four commonly used automatic metrics for the dialogue generation, the perplexity (\textbf{PPL}), unigram overlap (\textbf{F1}), \textbf{BLEU}, and distinct 1,2 (\textbf{Dist-1,2}).
\textbf{F1} and \textbf{BLEU} are commonly used to measure how similar the machine-generated responses is to referenced golden response \cite{Miller2017ParlAIAD,Papineni2002BleuAM}.
\textbf{Dist-1,2} measure the diversity of the generated responses \cite{li-etal-2016-diversity}.

\paragraph{Human Evaluations}
We perform human evaluation over the generated response.
Following \citet{song2021bob}, we consider three conventional criteria: fluency (\textbf{Flue.}), informativeness (\textbf{Info.}), and relevance (\textbf{Relv.}).
We recruit a team on Amazon Mechanical Turk consisting of several professional annotators, who are proficient in language tasks but know nothing about the models.\footnote{\url{https://www.mturk.com/}}
We sample $200$ instances for each model’s evaluation and each sample was evaluated by three people. 
Each criterion is rated on five scales, where 1, 3, and 5 indicate unacceptable, moderate, and perfect performance, respectively.
We report the average Fleiss’s kappa score \cite{fleiss1973equivalence} on \dataset{Reddit} and \dataset{Movie Dialogue}, 0.49 and 0.45 respectively, indicating annotators have reached moderate agreement. We also consider Sensibleness and Specificity Average (\textbf{SSA}), which evaluates two aspects of responses: making sense and being specific \cite{adiwardana2020towards}.

\begin{table*}[!t]
\scalebox{0.85}{
\begin{tabular}{l}
\toprule

\textbf{Speaker1}: Why do you get to decide who has something to offer ?  \\
\textbf{Speaker2}: He doesn't , he is entitled to his opinion , this is the internet and a forum discussion thread . \\ \hspace{1.7cm} People post their opinions not the truth . \\
\textbf{Baseline Generation}: \textcolor{red}{Why have you already voted to make sure you for yourself to support yourself ?} \\
\textbf{Key Evidence 1}: \textcolor{applegreen}{Everyone is entitled to an opinion , but those with experience in the area of discussion} \\ \hspace{2.7cm} \textcolor{applegreen}{usually have more pertinent and accurate opinions than others .} \\
\textbf{Key Evidence 2}: \textcolor{applegreen}{No you're entitled to your opinion . I'd just prefer an opinion that didn't contain a logical fallacy .} \\
\textbf{Our Generation}: \textcolor{blue}{I agree with you. People are entitled to their opinion . I just posted my own opinion . } \\
\textbf{Ground Truth}: I know , I was taking a round about way of trying to get him to questions his opinion . \\
\bottomrule
\end{tabular}}
\caption{\label{table: examples}
 Examples of responses generated by baseline and our approach based on \model{FiD}. 
}
\end{table*}

\begin{table*}
\small
\centering
\begin{tabular}{lcccccc|ccccc}
\toprule
 & & \multicolumn{5}{c|}{\dataset{Movie Dialogue} $\rightarrow$ \dataset{Reddit}} & \multicolumn{5}{c}{\dataset{Reddit} $\rightarrow$ \dataset{Movie Dialogue}}  \\
\cmidrule(lr){3-7}\cmidrule(lr){8-12}
 & & \textbf{PPL} & \textbf{F1} & \textbf{BLEU} & \textbf{Dist-1} & \textbf{Dist-2} & \textbf{PPL} & \textbf{F1} & \textbf{BLEU} & \textbf{Dist-1} & \textbf{Dist-2} \\ 
\midrule
\model{T5} & \model{baseline} & 29.2 &	5.3 &	3.9 &	95.6 &	96.2  & 33.0 	 &5.1 	 &3.6 	 &94.5 	 &95.9 \\
\multirow{3}{*}{\model{FiD} w. \model{SR}} 
 & \model{c2c} & 26.1 &	9.2 &	6.8 &	95.9 &	97.2   &27.3  &	8.8	 &6.7  &	95.8 	 & 96.7  \\
 & \model{c2r} & 26.2 &	9.1 &	6.6 &	95.2 &	96.6   &27.5  &	8.6  &	6.6  &	95.2  &	96.1  \\
 & \model{mix}  &\textbf{25.6}  &\textbf{9.8} &\textbf{7.3} &\textbf{96.4} &\textbf{98.1} & \textbf{26.8} 	&\textbf{9.5} &\textbf{7.1}  &\textbf{95.5} &\textbf{97.8}\\
\bottomrule
\end{tabular}
\caption{
\label{table:zero}
 Automatic evaluation results of zero-shot experiments over \dataset{Reddit} and \dataset{Movie Dialog} with 8 retrieved evidence passages. BERTScore is used to retrieve. The best results are in \textbf{bold}. 
}
\vspace{-0.2in}
\end{table*}

\subsection{Implementation and Setup}


As the context has a different number of turns, we use the latest utterance of dialogue context as the BERTScore query in practice, which can yield more consistent matching scores. Specifically, we compute F1$_{\text{BERT}}$ of context $c_i$ of current dialogue and the corresponding context of every evidence. We use \model{deberta-xlarge-mnli}\cite{he2020deberta} following the suggestion of authors.\footnote{\url{https://github.com/Tiiiger/bert_score}}
The filter is used in all retrieval setups except the baselines.

We perform an in-domain evaluation over the two datasets.
For each dataset, we adopt the proposed three self-retrieval (\model{SR}) method, \model{c2c}, \model{c2r}, and \model{mix}, comparing against the \model{GPT-2} and \model{FiD} baselines.
We experiments with different numbers of retrieval evidence passages (see \secref{sec:number_evidence}).
Note that \model{FiD} degenerates to a standard \model{T5} model without any evidence.
We retrain our model based on the pretrained checkpoint of \model{gpt-2},\footnote{\url{https://huggingface.co/gpt2/tree/main}} and T5 checkpoint for \model{FiD}.\footnote{\url{https://huggingface.co/t5-small/tree/main}}
We do model selection based on PPL over the validation set.

We additionally perform a zero-shot cross-domain evaluation for both datasets using \model{FiD}.\footnote{We ensure there is no overlap between the two datasets.}
In this setup, we only train our best in-domain \model{FiD} model on one dataset and then directly test on the other, while the retrieval set for inference is the training set of the target domain.
All other setups follow the in-domain experiments.

\subsection{Results}
\paragraph{In-domain}
Table~\ref{table: overview} reports the overall in-domain experimental results.
Overall, our self-retrieval methods achieve better performance consistently across almost all automatic and human evaluation metrics in terms of generating quality.
For generation diversities (\textbf{Dist-1 and Dist-2}), our SR can still have comparable performance with the strong baselines.
For both \model{GPT-2} and \model{FiD}, all three used matching strategies can improve the overall performance, and \model{mix} consistently outperforms the other two. 
Comparing with \model{GPT-2} and \model{FiD}, two baselines achieve similar performance, while when adding our retrieved evidences, we observed \model{FiD} based methods performance better, demonstrating the effectiveness of evidence-aware training of \model{FiD} in modeling multiple evidence passages.
We also illustrate the example generated by our approach and baselines in Table~\ref{table: examples}.
Above all, these results demonstrate that our approach could utilize more of the dialogue data without introducing more data compared with the baselines.

\paragraph{Zero-shot Cross-domain}
Table~\ref{table:zero} reports the results of zero-shot experiments using \model{FiD}.
Again, we find that our methods with evidence achieve better performance compared to the baselines without knowledge and \model{mix} performs the best.
This result indicates that our approach has good generalization and is robust to different datasets.

Overall, both in-domain and zero-shot results demonstrate our self-retrieval method can improve the performance of open-domain dialogue generation, and worth noting that our self-retrieval do not use any additional resources.
This indicates our methods can unleash more potential of the dialogue data compared with the vanilla training methods.

\begin{table*}
\small
\centering
\begin{tabular}{lcccccc|ccccc}
\toprule
\textbf{} & & \multicolumn{5}{c|}{\dataset{Reddit}} & \multicolumn{5}{c}{\dataset{Movie Dialog}}  \\
\cmidrule(lr){3-7}\cmidrule(lr){8-12}
\textbf{}  & & \textbf{PPL} & \textbf{F1} & \textbf{BLEU} & \textbf{Dist-1} & \textbf{Dist-2} & \textbf{PPL} & \textbf{F1} & \textbf{BLEU}  & \textbf{Dist-1} & \textbf{Dist-2}\\ 
\midrule
\model{GPT-2} & \model{baseline} &31.3 &5.3 &3.4 &65.4 &96.7 &25.6 &5.4 &3.3 &64.3 &96.0 \\
\multirow{5}{*}{SR} 
& p1 &28.3 	&6.9 &4.7 &74.5 &95.8 &22.8 	&6.8 	&4.6 	&71.3 	&94.2  \\
& p2 &27.9 	&7.1 &4.9 &74.2 &95.6  &22.5 	&7.1 	&4.8 	&71.6 &	94.8  \\
 & p4 &27.5	&7.4 &5.1 &75.1 &96.3  &22.1 	&7.3 	&5.0 &	72.8 	&95.3  \\
 & p8 &27.1 &7.8 &5.4 &76.1 &96.8 &21.6 &7.6 	&5.2 &	73.4 	&96.2  \\
 & p16 &\textbf{26.8} &\textbf{7.9} &\textbf{5.4} &\textbf{76.5} &\textbf{97.0} &\textbf{21.3} 	&\textbf{7.8} 	&\textbf{5.3} 	&\textbf{73.8}	&\textbf{96.5} \\
\midrule
\model{T5} & \model{baseline} & 25.5 &5.2 &3.7 &95.7 &96.3 	&20.5	&5.2	&3.7  &95.2 	&95.8 \\
\multirow{5}{*}{\model{FiD} w. \model{SR}}
& p1 &23.8 &9.5 &6.9 &93.7 &94.8 &19.1 	&9.0 	&6.3 &	94.6 	&95.7  \\
 & p2 &23.5 &9.8 &7.2 &94.1 &95.3  &18.7 	&9.4 	&6.7 	&94.4 	&95.5  \\
 & p4  &23.1 &10.1 &7.6 &94.6 &96.2  &18.2 	&9.8 	&7.2 	&94.9 	&96.3   \\
 & p8 &22.7 &10.4 &7.8 &95.6 &96.5 &17.9 	&10.1 	&7.5 	&95.3 	&96.9   \\
 & p16 &\textbf{22.4} &\textbf{10.6} &\textbf{7.9}  &\textbf{95.9}  &\textbf{98.2} &\textbf{17.7} &\textbf{10.3} &\textbf{7.6} &\textbf{95.5} &\textbf{97.0}\\
\bottomrule
\end{tabular}
\caption{
 Experimental results of different numbers of evidences used for generation using \dataset{Reddit} and \dataset{Movie Dialog}. p-$k$ indicates the number of evidence passages used for generation. The best results are in \textbf{bold}.}
\label{table: number}
\vspace{-0.1in}
\end{table*}

 \begin{table}[!t]
\small
\centering
\begin{tabular}{lcccc}
\toprule
\dataset{Reddit}  & & \textbf{PPL} & \textbf{F1} & \textbf{BLEU}   \\ 
\midrule
\model{GPT-2} & \model{baseline} &31.3 &5.3 &3.4  \\
\multirow{3}{*}{w. \model{SR}} 
 & \model{random} &31.4 	&5.4 	&3.4  	 \\
 & w/o \model{filter} &27.6	&7.2	&4.8  \\
 & w. \model{filter} &\textbf{27.1} &\textbf{7.8} &\textbf{5.4}  \\
\midrule
\model{\model{FiD}(\model{T5})} & \model{baseline}  & 25.5 &5.2 &3.7    \\
\multirow{3}{*}{\model{ w. \model{SR}}} 
 & \model{random} &25.7 	&5.2 	&3.6 	 \\
 & w/o \model{filter} &23.3 	&9.8	&7.2 	  \\
 & w. \model{filter} &\textbf{22.7} &\textbf{10.4} &\textbf{7.8}  \\

\bottomrule
\end{tabular}
\caption{Effectiveness of the Filter.}
\label{tab:filter}
\end{table}

\subsection{Analysis}
\label{sec:analysis}

\paragraph{Retrieval Methods}
Table~\ref{table: overview} shows the experimental results of different retrieval methods. We find that both methods achieve better results compared to baseline, which shows the generality of our self-retrieval method. We can also find that BERTScore performs better than BM25,\footnote{We only report the mix results for BM25. Refer to the appendix for full results.} which indicates that BERTScore could be used to get better retrieval evidences.

\paragraph{Retrieval Strategies}

Table~\ref{table: overview} also shows the experimental results of different retrieval strategies.
We find that \model{mix} perform better than context-to-context retrieval (\model{c2c}) and context-to-response retrieval (\model{c2r}), and the latter two methods show no significant difference.
We thought that both \model{c2c} and \model{c2r} can retrieve useful evidences while from different aspects.
And thus mixing them can yield more useful informative and relevant evidences and better performance as well.

\paragraph{Effectiveness of the Filter}
\label{sec:filter}
\tabref{tab:filter} shows the ablation study without using the filter during the retrieval step on \dataset{Reddit}.
Here the finding is that experiment with the filter (w. \model{filter}), has better performance than experiments without it (w/o \model{filter}), as well as a setup using random evidences (\model{random}).
These show that noisy evidences give no assistance, or even harm, to the model and that the necessity of discarding low-relevant evidence in our method.

\paragraph{Number of Retrieved Evidences}
\label{sec:number_evidence}
We also carried out experiments with a different number of retrieved evidences.
\tabref{table: number} reports the experimental results of using $k$ evidences~(p-$k$) for generation.
We observe that experiment using more retrieved evidences (p$16$) performs better than experiments with fewer retrieved evidences (i.e. p$1$, p$2$, p$4$, p$8$).
While the performance gap is getting smaller when increasing the evidence numbers.
Considering the trade-off between efficiency and performance, we report results using $8$ evidence as our main results, which is considered to be good enough.
These results indicate that we can use more retrieved evidences to obtain better experimental results. In addition, supporting more information is significant for the generative model.

\begin{figure*}[!t]
\centering
\begin{subfigure}[b]{0.6\columnwidth}
\centering
\includegraphics[width=1\textwidth]{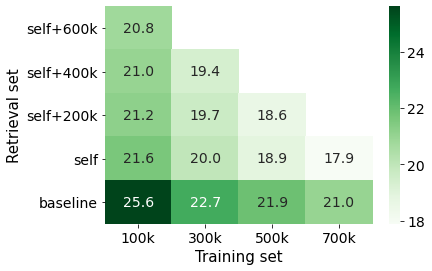}
\caption{PPL on \model{GPT-2}}
\end{subfigure}
\hfill
\begin{subfigure}[b]{0.6\columnwidth}
\centering
\includegraphics[width=1\textwidth]{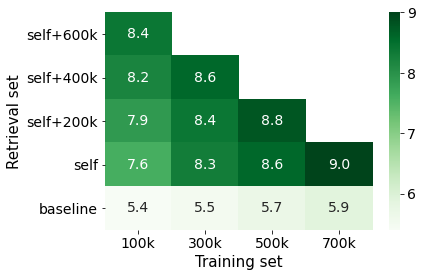}
\caption{F1 on \model{GPT-2}}
\end{subfigure}
\hfill
\begin{subfigure}[b]{0.6\columnwidth}
\centering
\includegraphics[width=1\textwidth]{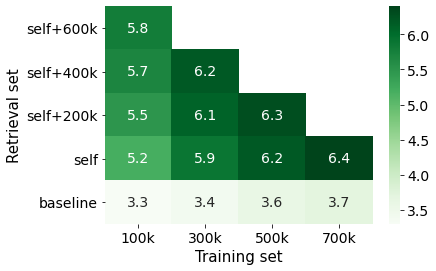}
\caption{BLEU on \model{GPT-2}}
\end{subfigure}

\begin{subfigure}[b]{0.6\columnwidth}
\centering
\includegraphics[width=1\textwidth]{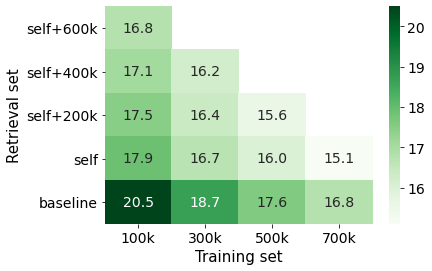}
\caption{PPL on \model{FiD}}
\end{subfigure}
\hfill
\begin{subfigure}[b]{0.6\columnwidth}
\centering
\includegraphics[width=1\textwidth]{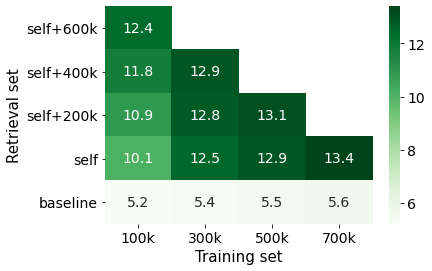}
\caption{F1 on \model{FiD}}
\end{subfigure}
\hfill
\begin{subfigure}[b]{0.6\columnwidth}
\centering
\includegraphics[width=1\textwidth]{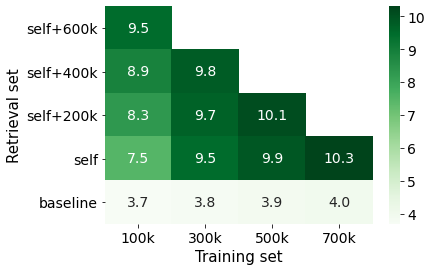}
\caption{BLEU on \model{FiD}}
\end{subfigure}
\caption{Results of different sizes of training set and retrieval set on the \dataset{Movie Dialog} with 8 retrieved evidences. ``Self'' indicates the training set used for self-retrieval and ``+'' means adding extra data for retrieval.}
\label{figure: heatmap}
\end{figure*}

\begin{figure*}[!t]
\centering
\begin{subfigure}[b]{1\columnwidth}
\centering
\includegraphics[width=0.9\textwidth]{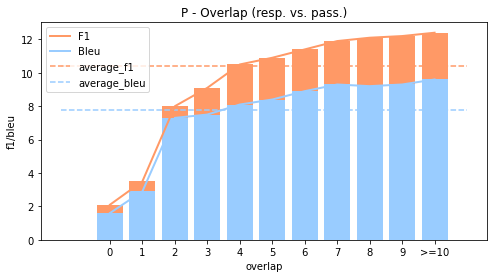}
\caption{$\max$ setup over overlaps with bins = $\{0,\cdots,9,\geq10\}$}
\end{subfigure}
\hfill
\begin{subfigure}[b]{1\columnwidth}
\centering
\includegraphics[width=0.9\textwidth]{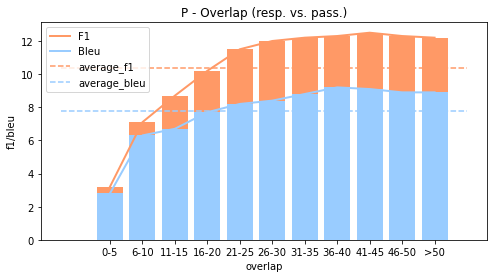}
\caption{$\operatorname{sum}$ setup over overlaps using bin size = 5}
\end{subfigure}
\caption{Performance by different overlaps between evidences and ground-truth responses over \dataset{Reddit}.} 
\label{figure: overlap}
\end{figure*}

\paragraph{Self-retrieval vs. Extra Evidences}
\label{sec:extra}
We made the retrieval set exactly the same as the training set, denoted as the ``self-retrieval (\model{SR})'' setup.
One natural question is \textit{can we use extra data for retrieval set?}
To further understand this question and to validate the usefulness of our method, we carried out experiments with different sizes of the training set and retrieval set.
Specifically, we experiment with additional setups by enlarging the retrieval sets, i.e. +200k, +400k, +600k, where ``+'' means extra data for retrieval sets, and we also adopt baselines with different training sizes of 100k, 300k, 500k, 700k (denoted before ``+'').\footnote{Due to data size limitation, we did not occupy all setups.}

Figure~\ref{figure: heatmap} shows the experimental results.\footnote{We also report a detailed results using (100+600k) setup in Appendix~\ref{sec:full_extra}.}
We observe that experiments with larger retrieval sets achieve better results than those with small retrieval sets across different training sizes.
We believe larger retrieval sets can introduce more relevant evidences, which brings performance gain for the model. Another interesting finding is that adding extra data for retrieval (100+600k, 300+400k, 500+200k) can outperform the baselines (700k) with extra data added via direct training.
Also, under the same amount of total data (700k), leveraging more data for retrieval (100+600k, 300+400k, 500+200k) has approaching performance with the self-retrieval with full data (self, 700k).
It indicates that our methods can increase the usage of the training data only in a retrieval way without directly training these responses, and our method has good generalization over the retrieval evidences.

\paragraph{Relevance of Evidence and Ground-truth}
\label{sec:relev}
To further study how our methods make sense, we study \textit{how the relevance of the retrieved evidences and ground-truth response can influence the generation performance}.
For each instance $(c_i, r_i)$ which used $n$ retrieval evidences $\set{E}_i^{\model{mix}} = \{e_1, e_2,...,e_{n} \}$, we compute the number of overlapped words between the ground-truth $r_i$ and each retrieved evidence.
We study two setups by computing the overall $\operatorname{overlap}(\set{E}, r_i)$ using $\max$ and $\operatorname{sum}$ over the individual overlaps.

Figure~\ref{figure: overlap} shows the results of these two setups.
We observed that higher overlap leads to better performance.
It indicates that high relevant retrieval evidences can help to generate better responses and low relevant knowledge are harmful, which is consistent with the findings in \secref{sec:filter}.
Also, there are low-relevant evidences left in the retrieval step, which indicates that open-domain dialogue generation is still a difficult task, and better retrieval methods are required to further improve our generation performance.

\section{Conclusion}
In this paper, we propose a self-retrieval training framework for open-domain dialogue generation.
Different from other knowledge-intensive tasks, our framework only retrieves relevant dialogue instances from the training data (which can be extended to a retrieval set) without the need to train them in the generation model.
It is significant that we demonstrate that traditional training baselines underutilize the training data and our method can utilize more potential of data.
We show that our method improves the robustness and generality of generative models as well as generate proper response for complicated human conversation.
We also find that BERTScore can be used for better evidence retrieval.
In future works, we would like to study better ways of evidence retrieval and evidence-aware training and we believe our approach can benefit to other NLP tasks, such as classification task.

\section*{Acknowledgements}
The authors would like to thank the anonymous reviewers for their constructive and insightful comments. This work was supported by the National Natural Science Foundation of China under Grants No. 61972290.



\bibliography{acl}
\bibliographystyle{acl}

\clearpage
\appendix

\section{Appendix}
\label{sec:appendix}

\subsection{Full results of Retrieving Extra data}
\label{sec:full_extra}
We present a full results of enlarging the retrieval set to (100+600k) for both \dataset{Reddit} and \dataset{Movie Dialogue}, shown in \tabref{table:large_res}.
The training set is 100k as the same as the self-retrieval setup in main results. BM25 is used to retrieve.

\begin{table}[!h]
\centering
\small
\begin{tabular}{llccc}
\toprule
 \midrule
 
\dataset{Reddit} & & \textbf{PPL}$\downarrow$ & \textbf{F1}$\uparrow$ & \textbf{BLEU}$\uparrow$ \\

\midrule
GPT-2 & \model{baseline}  &31.3 &5.3 &3.4  \\

\multirow{3}{*}{GPT-2 w. DR} & \model{c2c}  &28.0 	&6.2 	&3.8  \\
     & \model{c2r}  &28.2 	&6.0 	&3.7 	\\
     & \model{mix}  &\textbf{26.9} 	&\textbf{6.8} 	&\textbf{4.3} 	\\
\midrule
T5  & \model{baseline} & 25.5 	&5.2	&3.7 \\
\multirow{3}{*}{\model{FiD} w. DR} & \model{c2c}  &23.6	&9.6 	&7.2 	 \\
 & \model{c2r}   &23.8	&9.4 	&7.1 	\\
 & \model{mix}   &\textbf{21.9}	&\textbf{12.0} 	&\textbf{9.0}  \\
\midrule \midrule

\dataset{Movie} & & \textbf{PPL}$\downarrow$ & \textbf{F1}$\uparrow$ & \textbf{BLEU}$\uparrow$  \\

\midrule
GPT2  &\model{baseline} & 25.6 &5.4 &3.3 \\

\multirow{3}{*}{GPT-2 w. DR} & \model{c2c}  &22.5 &6.0 &3.7  \\
 & \model{c2r}  &22.6 &5.9 &3.5  \\
 & \model{mix}  &\textbf{21.7} 	&\textbf{7.3}	&\textbf{4.7} 	 \\
\midrule
T5  &\model{baseline}	&20.5	&5.2	&3.7   \\
\multirow{3}{*}{\model{FiD} w. DR} & \model{c2c}  &19.2	&9.1	&6.9  \\
 & \model{c2r}  &19.4	&9.0	&6.7  \\
 & \model{mix}  &\textbf{17.7}	&\textbf{11.5}	&\textbf{8.5}   \\
\bottomrule
\end{tabular}
\caption{
 Automatic evaluations of the in-domain setups on the \dataset{Reddit} and \dataset{Movie Dialog} datasets with 8 evidences for retrieval. The best results are in \textbf{bold}. 
}
\label{table:large_res}
\end{table}

\subsection{Full results of Self-Retrieval}
\label{sec:full_results}

\begin{table*}[!ht]
\centering
\small
\begin{tabular}{lcccccc|cccc}
\toprule
 & & \multicolumn{5}{c}{Automatic Metrics} & \multicolumn{3}{c}{Human Evaluation}\\
 \cmidrule(lr){3-7} \cmidrule(lr){8-11}
 
\dataset{Reddit} & & \textbf{PPL}$\downarrow$ & \textbf{F1}$\uparrow$ & \textbf{BLEU}$\uparrow$ & \textbf{Dist-1}$\uparrow$ & \textbf{Dist-2}$\uparrow$ & \textbf{Flue}$\uparrow$ & \textbf{Info}$\uparrow$ & \textbf{Relv}$\uparrow$ & \textbf{SSA}$\uparrow$\\

\midrule
\model{GPT-2} & \model{baseline} & 31.3 &5.3 &3.4  &65.4 &96.7  &3.0  &2.9 	&2.8  &46\%\\
\multirow{6}{*}{w. \model{SR}}
& BM25 \model{c2c}  &29.4 &6.1 	&3.8 &69.3 	&95.6 &3.2  &3.0 	 &3.1 &49\%\\
& BM25 \model{c2r}  &29.7 &6.0 	&3.6 &68.4 	&95.3 &3.2  &3.1 	 &3.1 &50\%\\
& BM25 \model{mix}  &28.1 &6.6 	&4.2 &73.5 	&\textbf{98.2} &3.4  &3.3 	 &3.4 &51\%\\
&BERTScore \model{c2c}  &27.7 &7.2 	&4.8 	&75.2 	&96.1 &3.4  &3.4 	 &3.4 &54\%\\
  &BERTScore \model{c2r}  &27.9 	&7.0 	&4.7 	&75.0	&96.4  &3.3 	 &3.4 	 &3.3 &53\%\\
   &BERTScore \model{mix}  &\textbf{27.1} &\textbf{7.8}	&\textbf{5.4}	&\textbf{76.1}	&96.8  	&\textbf{3.5}	&\textbf{3.4}	&\textbf{3.5} &\textbf{55\%}\\
\midrule

\model{T5}  &\model{baseline} & 25.5 	&5.2	&3.7 	&\textbf{95.7}	&96.3 &3.1 &3.0 &3.1 &48\%\\
\multirow{6}{*}{\model{FiD} w. \model{SR}} 
& BM25 \model{c2c}  &25.0 &8.0 	&5.9 &91.2 	&93.8 &3.3  &3.2 	 &3.3 &51\%\\
& BM25 \model{c2r}  &25.2 &7.9 	&5.7 &90.4 	&92.3 &3.3  &3.2 	 &3.2 &50\%\\
& BM25 \model{mix}  &23.8 &9.5 	&6.9 	&95.3 	&\textbf{97.2} &3.5 &3.4  &3.5 &52\%\\
 &BERTScore \model{c2c}  &23.3	&9.9 	&7.3	&94.3 	&94.7 &3.5 &3.5 &3.4 &54\%\\
 &BERTScore \model{c2r}   &23.4	&9.8 	&7.2 	&94.0 	&94.4 &3.5 &3.4 &3.4 &54\%\\
 &BERTScore \model{mix}   &\textbf{22.7}	&\textbf{10.4}	&\textbf{7.8} 	&95.6	&96.5 &\textbf{3.6} &\textbf{3.5} &\textbf{3.5} &\textbf{56\%}\\
\midrule \midrule

\dataset{Movie} & & \textbf{PPL} & \textbf{F1} & \textbf{BLEU} & \textbf{Dist-1} & \textbf{Dist-2} & \textbf{Flue} & \textbf{Info} & \textbf{Relv} & \textbf{SSA}\\

\midrule
\model{GPT-2}  & \model{baseline} & 25.6 &5.4 &3.3 &64.3 &96.0 &3.0 &2.9 &2.8 &47\%\\
\multirow{6}{*}{w. \model{SR}} 

& BM25 \model{c2c}  &23.5 &6.1 	&3.8 &66.9 	&93.9 &3.2  &3.1 	 &3.1 &51\%\\
& BM25 \model{c2r}  &23.5 &6.0 	&3.7 &67.8	&92.7 &3.2  &3.0 	 &3.1 &50\%\\
& BM25 \model{mix}  &22.7 &6.7 	&4.2 	&71.4 	&96.1 &3.4 &3.3  &3.3 &52\%\\
 &BERTScore \model{c2c}  &22.1 &7.1 &4.7 &71.7 &94.9 &3.4 	&3.4 	&3.3 &53\%\\
 &BERTScore \model{c2r}  &22.3 &7.0 &4.7 &72.0 &94.3  &3.3 	&3.4 	&3.3 &53\%\\
 &BERTScore \model{mix}  &\textbf{21.6} 	&\textbf{7.6}	&\textbf{5.2}	&\textbf{73.4}	&\textbf{96.2} &\textbf{3.5}	&\textbf{3.4} 	&\textbf{3.4} &\textbf{55\%}\\
\midrule
\model{T5}  & \model{baseline}	&20.5	&5.2	&3.7  &95.2 	&95.8 &3.1 	&2.9 &	2.9  &48\%\\
\multirow{6}{*}{\model{FiD} w. \model{SR}}
& BM25 \model{c2c}  &20.1 &7.7 	&5.5 &92.3	&94.1 &3.3  &3.2 	 &3.2 &52\%\\
& BM25 \model{c2r}  &20.2 &7.7 	&5.4 &91.7 	&93.6 &3.3  &3.1 	 &3.2 &51\%\\
 & BM25 \model{mix}  &18.9	&9.2 &6.6  &94.9 &96.8 &3.6 &3.5 &3.6 &53\%\\
 &BERTScore \model{c2c}  &18.4	&9.5 &7.0  &94.4 &95.6 &3.6 &3.6 &3.5 &55\%\\
 &BERTScore \model{c2r}  &18.5	&9.4	&6.8  &93.8 	&94.9 &3.5 	&3.5 	&3.6 &54\%\\
 &BERTScore \model{mix}  &\textbf{17.9}	&\textbf{10.1}	&\textbf{7.5}  &\textbf{95.3}	&\textbf{96.9} &\textbf{3.7}	&\textbf{3.7}	&\textbf{3.6} &\textbf{57\%}\\
\bottomrule
\end{tabular}
\caption{
 Automatic and human evaluation of the in-domain setups over \dataset{Reddit} and \dataset{Movie Dialog}, using 8 evidences passages. \model{GPT-2} and \model{T5} are baselines.  ``w. \model{SR}'' (with self-retrieval) indicate our methods. The best results are in \textbf{bold}. 
}
\label{table: overview_all}
\end{table*}

\end{document}